\documentclass{article}
\pdfoutput=1

\PassOptionsToPackage{numbers, compress}{natbib}
\usepackage[preprint]{nips_2018}

\usepackage[utf8]{inputenc} 
\usepackage[T1]{fontenc}    
\usepackage{url}            
\usepackage{booktabs}       
\usepackage{amsfonts}       
\usepackage{nicefrac}       
\usepackage{microtype}      
\usepackage{color}
\usepackage[]{graphicx}
\usepackage{amsmath}
\usepackage{xcolor}
\usepackage{paralist}
\usepackage{enumerate}
\usepackage[]{subcaption}
\usepackage{listings}
\usepackage{courier}

\mathchardef\mhyphen="2D

\title{Relational recurrent neural networks}

\author{
Adam Santoro*$^{\alpha}$, Ryan Faulkner*$^{\alpha}$, David Raposo*$^{\alpha}$, Jack Rae$^{\alpha \beta}$, Mike Chrzanowski$^{\alpha}$,\\
\textbf{Th\'eophane Weber$^{\alpha}$, Daan Wierstra$^{\alpha}$, Oriol Vinyals$^{\alpha}$, Razvan Pascanu$^{\alpha}$, Timothy Lillicrap$^{\alpha \beta}$}\\
\\
*Equal Contribution\\
\\
$^{\alpha}$DeepMind\\
London, United Kingdom\\
\\
$^{\beta}$CoMPLEX, Computer Science, University College London\\
London, United Kingdom\\
\\
\small\texttt{\{adamsantoro; rfaulk; draposo; jwrae; chrzanowskim;}\\
\small\texttt{theophane; weirstra; vinyals; razp; countzero\}@google.com}
}

\DeclareFixedFont{\ttb}{T1}{txtt}{bx}{n}{8}
\begin{document}

\maketitle

\begin{abstract}
Memory-based neural networks model temporal data by leveraging an ability to remember information for long periods. It is unclear, however, whether they also have an ability to perform complex relational reasoning with the information they remember. Here, we first confirm our intuitions that standard memory architectures may struggle at tasks that heavily involve an understanding of the ways in which entities are connected -- i.e., tasks involving relational reasoning. We then improve upon these deficits by using a new memory module -- a \textit{Relational Memory Core} (RMC) -- which employs multi-head dot product attention to allow memories to interact. Finally, we test the RMC on a suite of tasks that may profit from more capable relational reasoning across sequential information, and show large gains in RL domains (e.g. Mini PacMan), program evaluation, and language modeling, achieving state-of-the-art results on the WikiText-103, Project Gutenberg, and GigaWord datasets.
\end{abstract}

\section{Introduction}

Humans use sophisticated memory systems to access and reason about important information regardless of when it was initially perceived \citep{schacter1994memory, knowlton2012neurocomputational}. In neural network research many successful approaches to modeling sequential data also use memory systems, such as LSTMs \citep{hochreiter1998lstm} and memory-augmented neural networks generally \citep{graves2014neural,graves2016hybrid,santoro2016meta,sukhbaatar2015end}. Bolstered by augmented memory capacities, bounded computational costs over time, and an ability to deal with vanishing gradients, these networks learn to correlate events across time to be proficient at \textit{storing} and \textit{retrieving} information.

Here we propose that it is fruitful to consider \textit{memory interactions} along with storage and retrieval. Although current models can learn to compartmentalize and relate distributed, vectorized memories, they are not biased towards doing so explicitly. We hypothesize that such a bias may allow a model to better understand how memories are related, and hence may give it a better capacity for relational reasoning over time. We begin by demonstrating that current models do indeed struggle in this domain by developing a toy task to stress relational reasoning of sequential information. Using a new \textit{Relational Memory Core} (RMC), which uses multi-head dot product attention to allow memories to interact with each other, we solve and analyze this toy problem. We then apply the RMC to a suite of tasks that may profit from more explicit memory-memory interactions, and hence, a potentially increased capacity for relational reasoning across time: partially observed reinforcement learning tasks, program evaluation, and language modeling on the Wikitext-103, Project Gutenberg, and GigaWord datasets.

\section{Relational reasoning}

We take relational reasoning to be the process of understanding the ways in which entities are connected and using this understanding to accomplish some higher order goal \citep{waltz1999system}. For example, consider sorting the distances of various trees to a park bench: the \textit{relations} (distances) between the \textit{entities} (trees and bench) are compared and contrasted to produce the solution, which could not be reached if one reasoned about the properties (positions) of each individual entity in isolation.

Since we can often quite fluidly define what constitutes an ``entity'' or a ``relation'', one can imagine a spectrum of neural network inductive biases that can be cast in the language of relational reasoning \footnote{Indeed, in the broadest sense any multivariable function must be considered ``relational.''}. For example, a convolutional kernel can be said to compute a relation (linear combination) of the entities (pixels) within a receptive field. Some previous approaches make the relational inductive bias more explicit: in message passing neural networks \citep[e.g.][]{gilmer2017neural,scarselli2009graph, LiTBZ15, battaglia2016interaction}, the nodes comprise the entities and relations are computed using learnable functions applied to nodes connected with an edge, or sometimes reducing the relational function to a weighted sum of the source entities \citep[e.g.][]{kipf2016semi, velickovic2018graph}. In Relation Networks \citep{santoro2017simple,raposo2017discovering,hu2017relation} entities are obtained by exploiting spatial locality in the input image, and the model focuses on computing binary relations between each entity pair. Even further, some approaches emphasize that more capable reasoning may be possible by employing simple computational principles; by recognizing that relations might not always be tied to proximity in space, \textit{non-local} computations may be better able to capture the relations between entities located far away from each other \citep{wang2017non,liu2018non}. 

In the temporal domain relational reasoning could comprise a capacity to compare and contrast information seen at different points in time \cite{pavez2018working}. Here, attention mechanisms \citep[e.g.][]{BahdanauCB14,vaswani2017attention} implicitly perform some form of relational reasoning; if previous hidden states are interpreted as entities, then computing a weighted sum of entities using attention helps to remove the \textit{locality bias} present in vanilla RNNs, allowing embeddings to be better related using content rather than proximity. 

Since our current architectures solve complicated temporal tasks they must have some capacity for temporal relational reasoning. However, it is unclear whether their inductive biases are limiting, and whether these limitations can be exposed with tasks demanding particular types of temporal relational reasoning. For example, memory-augmented neural networks \citep{graves2014neural,graves2016hybrid,santoro2016meta,sukhbaatar2015end} solve a compartmentalization problem with a slot-based memory matrix, but may have a harder time allowing memories to interact, or relate, with one another once they are encoded. LSTMs \citep{hochreiter1998lstm, Graves13}, on the other hand, pack all information into a common hidden memory vector, potentially making compartmentalization and relational reasoning more difficult. 
\section{Model}
Our guiding design principle is to provide an architectural backbone upon which a model can learn to compartmentalize information, and learn to compute interactions between compartmentalized information. To accomplish this we assemble building blocks from LSTMs, memory-augmented neural networks, and non-local networks (in particular, the Transformer seq2seq model \cite{vaswani2017attention}). Similar to memory-augmented architectures we consider a fixed set of memory slots; however, we allow for interactions \textit{between} memory slots using an attention mechanism. As we will describe, in contrast to previous work we apply attention between memories at a single time step, and not across all previous representations computed from all previous observations.

\begin{figure}
    \centering
    \includegraphics[width=1\textwidth]{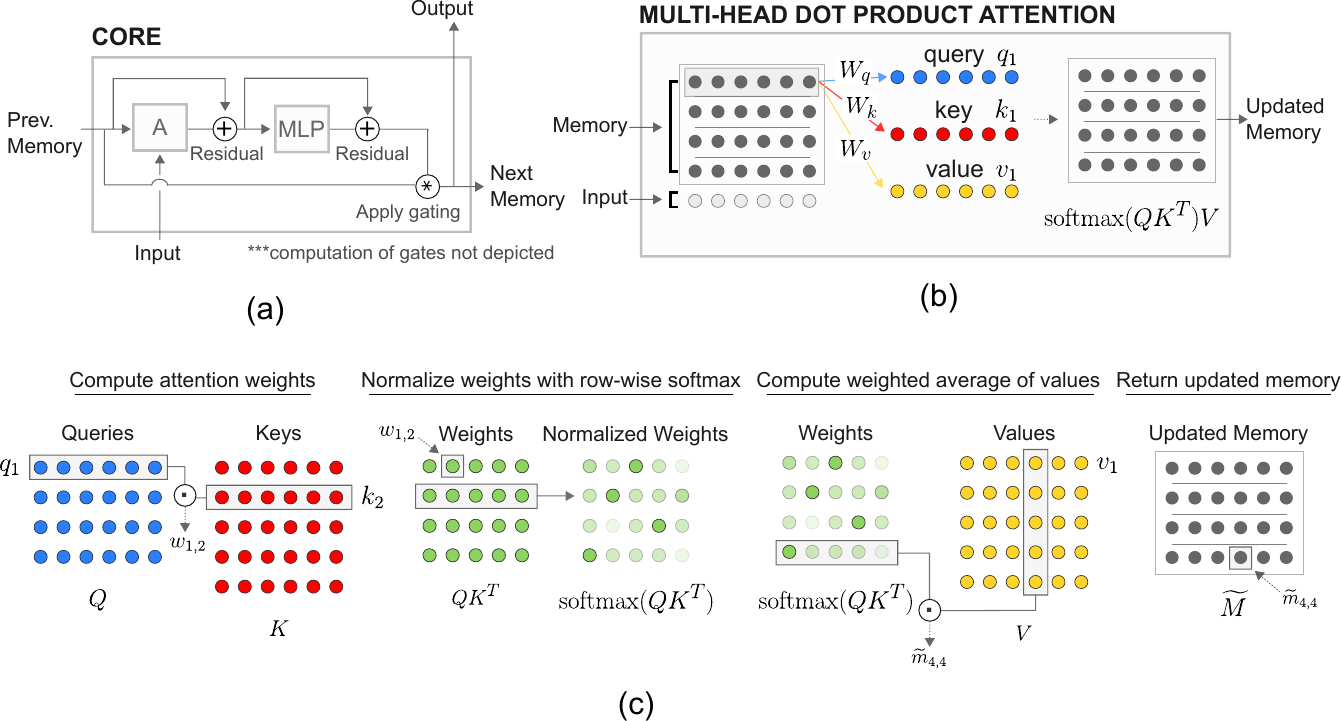}
    \caption{\textbf{Relational Memory Core}. (a) The RMC receives a previous memory matrix and input vector as inputs, which are passed to the MHDPA module labeled with an ``A''. (b). Linear projections are computed for each memory slot, and input vector, using row-wise shared weights $W^q$ for the queries, $W^k$ for the keys, and $W^v$ for the values. (c) The queries, keys, and values are then compiled into matrices and $\text{softmax}(QK^T)V$ is computed. The output of this computation is a new memory where information is blended across memories based on their attention weights. An MLP is applied row-wise to the output of the MHDPA module (a), and the resultant memory matrix is gated, and passed on as the core output or next memory state.}
    \label{fig:model}
\end{figure}

\subsection{Allowing memories to interact using multi-head dot product attention}

We will first assume that we do not need to consider memory encoding; that is, that we already have some stored memories in matrix $M$, with row-wise compartmentalized memories $m_i$. To allow memories to interact we employ \textit{multi-head dot product attention} (MHDPA) \citep{vaswani2017attention}, also known as \textit{self-attention}. Using MHDPA, each memory will attend over all of the other memories, and will update its content based on the attended information.

First, a simple linear projection is used to construct queries ($Q = MW^q$), keys ($K = MW^k$), and values ($V = MW^v$) for each memory (i.e. row $m_i$) in matrix $M$. Next, we use the queries, $Q$, to perform a scaled dot-product attention over the keys, $K$. The returned scalars can be put through a softmax-function to produce a set of weights, which can then be used to return a weighted average of values from $V$ as $A(Q, K, V) = \text{softmax} \left( \frac{QK^T}{\sqrt{d_k}} \right) V $, where $d_k$ is the dimensionality of the key vectors used as a scaling factor. Equivalently:
\begin{align}
    A_{\theta}(M) = \text{softmax} \left( \frac{MW^q(MW^k)^T}{\sqrt{d_k}} \right) MW^v \label{eq:attention}, \text{ where } \theta=(W^q, W^k, W^v)
\end{align}

The output of $A_{\theta}(M)$, which we will denote as $\widetilde{M}$, is a matrix with the same dimensionality as $M$. $\widetilde{M}$ can be interpreted as a proposed update to $M$, with each $\widetilde{m}_i$ comprising information from memories $m_{j}$. Thus, in one step of attention each memory is updated with information originating from other memories, and it is up to the model to learn (via parameters $W^q$, $W^k$, and $W^v$) how to shuttle information from memory to memory.

As implied by the name, MHDPA uses multiple heads. We implement this producing $h$ sets of queries, keys, and values, using unique parameters to compute a linear projection from the original memory for each head $h$. We then independently apply an attention operation for each head. For example, if $M$ is an $N \times F$ dimensional matrix and we employ two attention heads, then we compute $\widetilde{M^1} = A_{\theta}(M)$ and $\widetilde{M^2} = A_{\phi}(M)$, where $\widetilde{M^1}$ and $\widetilde{M^2}$ are $N \times F/2$ matrices, $\theta$ and $\phi$ denote unique parameters for the linear projections to produce the queries, keys, and values, and $\widetilde{M} = [\widetilde{M^1}: \widetilde{M^2}]$, where $[:]$ denotes column-wise concatenation. Intuitively, heads could be useful for letting a memory share different information, to different targets, using each head.

\subsection{Encoding new memories}
We assumed that we already had a matrix of memories $M$. Of course, memories instead need to be encoded as new inputs are received. Suppose then that $M$ is some randomly initialised memory. We can efficiently incorporate new information $x$ into $M$ with a simple modification to equation \ref{eq:attention}:
\begin{align}
\widetilde{M} = \text{softmax} \left( \frac{MW^q([M;x]W^k)^T}{\sqrt{d^k}} \right) [M;x]W^v, \label{eq:attention_memory}
\end{align}
where we use $[M;x]$ to denote the row-wise concatenation of $M$ and $x$. Since we use $[M;x]$ when computing the keys and values, and only $M$ when computing the queries, $\widetilde{M}$ is a matrix with same dimensionality as $M$. Thus, equation \ref{eq:attention_memory} is a memory-size preserving attention operation that includes attention over the memories and the new observations. Notably, we use the same attention operation to efficiently compute memory interactions and to incorporate new information. 

We also note the possible utility of this operation when the memory consists of a single vector rather than a matrix. In this case the model may learn to pick and choose which information from the input should be written into the vector memory state by learning how to attend to the input, conditioned on what is contained in the memory already. This is possible in LSTMs via the gates, though at a different granularity. We return to this idea, and the possible compartmentalization that can occur via the heads even in the single-memory-slot case, in the discussion. 

\subsection{Introducing recurrence and embedding into an LSTM}
Suppose we have a temporal dimension with new observations at each timestep, $x_t$. Since $M$ and $\widetilde{M}$ are the same dimensionality, we can naively introduce recurrence by first randomly initialising $M$, and then updating it with $\widetilde{M}$ at each timestep. We chose to do this by embedding this update into an LSTM. Suppose memory matrix $M$ can be interpreted as a matrix of cell states, usually denoted as $C$, for a 2-dimensional LSTM. We can make the operations of individual memories $m_i$ nearly identical to those in a normal LSTM cell state as follows (subscripts are overloaded to denote the row from a matrix, and timestep; e.g., $m_{i, t}$ is the $i^{th}$ row from $M$ at time $t$).
\begin{align}
    s_{i, t} &= (h_{i, t-1}, m_{i, t-1}) \\
    f_{i, t} &= W^f x_t + U^f h_{i, t-1} + b^f \\
    i_{i, t} &= W^i x_t + U^i h_{i, t-1} + b^i \\
    o_{i, t} &= W^o x_t + U^o h_{i, t-1} + b^o \\
    m_{i, t} &= \sigma(f_{i,t} + \tilde{b}^{f}) \circ m_{i, t-1} + \sigma(i_{i, t}) \circ \underbrace{{g_{\psi}(\widetilde{m}_{i,t})}} \\
    h_{i, t} &= \sigma(o_{i, t}) \circ \text{tanh}(m_{i, t}) \\
    s_{i, t+1} &= (m_{i, t}, h_{i, t})
\end{align}

The underbrace denotes the modification to a standard LSTM. In practice we did not find output gates necessary -- please see the url in the footnote for our Tensorflow implementation of this model in the Sonnet library \footnote{\url{https://github.com/deepmind/sonnet/blob/master/sonnet/python/modules/relational_memory.py}}, and for the exact formulation we used, including our choice for the $g_{\psi}$ function (briefly, we found a row/memory-wise MLP with layer normalisation to work best). There is also an interesting opportunity to introduce a different kind of gating, which we call `memory' gating, which resembles previous gating ideas \cite{gers1999learning, hochreiter1998lstm}. Instead of producing scalar gates for each individual unit (`unit' gating), we can produce scalar gates for each memory row by converting $W^f$, $W^i$, $W^o$, $U^f$, $U^i$, and $U^o$ from weight matrices into weight vectors, and by replacing the element-wise product in the gating equations with scalar-vector multiplication. 

Since parameters $W^f$, $W^i$, $W^o$, $U^f$, $U^i$, $U^o$, and $\psi$ are shared for each $m_i$, we can modify the number of memories without affecting the number of parameters. Thus, tuning the number of memories and the size of each memory can be used to balance the overall storage capacity (equal to the total number of units, or elements, in $M$) and the number of parameters (proportional to the dimensionality of $m_i$). We find in our experiments that some tasks require more, but not necessarily larger, memories, and others such as language modeling require fewer, larger memories.

Thus, we have a number of tune-able parameters: the number of memories, the size of each memory, the number of attention heads, the number of steps of attention, the gating method, and the post-attention processor $g_{\psi}$. In the appendix we list the exact configurations for each task.
\section{Experiments}
Here we briefly outline the tasks on which we applied the RMC, and direct the reader to the appendix for full details on each task and details on hyperparameter settings for the model. 

\begin{figure}
    \centering
    \includegraphics[width=1\textwidth]{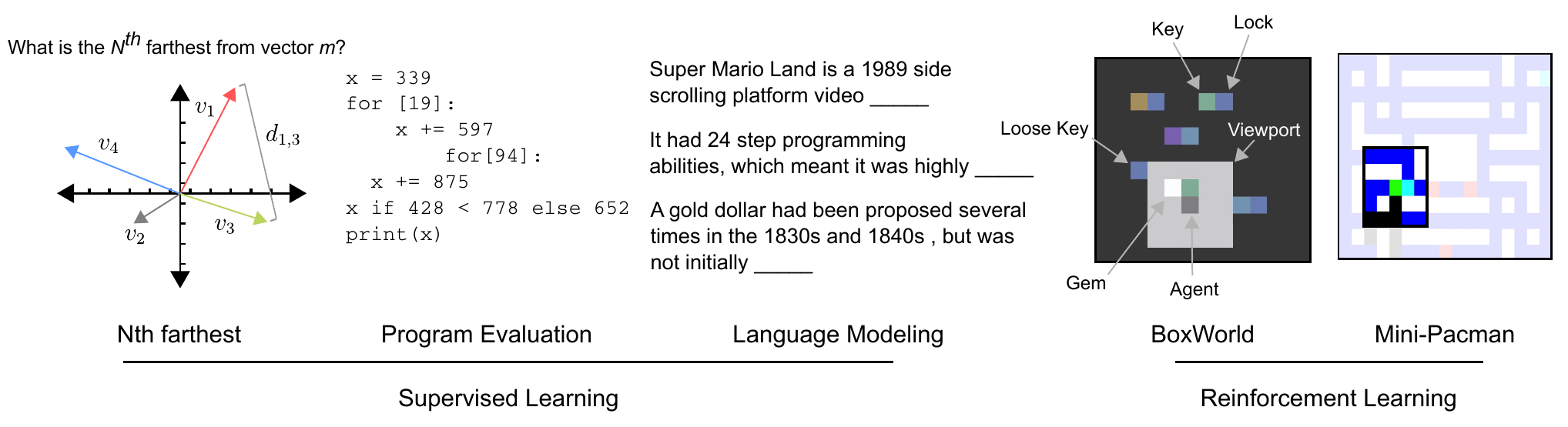}
    \caption{\textbf{Tasks}. We tested the RMC on a suite of supervised and reinforcement learning tasks. Notable are the $N^{th}$ Farthest toy task and language modeling. In the former, the solution requires explicit relational reasoning since the model must sort \textit{distance relations} between vectors, and not the vectors themselves. The latter tests the model on a large quantity of natural data and allows us to compare performance to well-tuned models.}
    \label{fig:tasks}
\end{figure}

\subsection{Illustrative supervised tasks}
\paragraph{$N^{th}$ Farthest}
The $N^{th}$ Farthest task is designed to stress a capacity for relational reasoning across time. Inputs are a sequence of randomly sampled vectors, and targets are answers to a question of the form: ``What is the $n^{th}$ farthest vector (in Euclidean distance) from vector $m$?'', where the vector values, their IDs, $n$, and $m$ are randomly sampled per sequence. It is not enough to simply encode and retrieve information as in a copy task. Instead, a model must compute all pairwise distance relations to the reference vector $m$, which might also lie in memory, or might not have even been provided as input yet. It must then implicitly sort these distances to produce the answer. We emphasize that the model must sort \textit{distance relations} between vectors, and not the vectors themselves.

\paragraph{Program Evaluation}
The \textit{Learning to Execute} (\textbf{LTE}) dataset \cite{zaremba2014lte} consists of algorithmic snippets from a Turing complete programming language of pseudo-code, and is broken down into three categories: \textit{addition}, \textit{control}, and \textit{full program}. Inputs are a sequence of characters over an alphanumeric vocabulary representing such snippets, and the target is a numeric sequence of characters that is the execution output for the given programmatic input. Given that the snippets involve symbolic manipulation of variables, we felt it could strain a model's capacity for relational reasoning; since symbolic operators can be interpreted as defining a relation over the operands, successful learning could reflect an understanding of this relation. To also assess model performance on classical sequence tasks we also evaluated on \textit{memorization tasks}, in which the output is simply a permuted form of the input rather than an evaluation from a set of operational instructions.  See the appendix for further experimental details.

\subsection{Reinforcement learning}

\paragraph{Mini Pacman with viewport} We follow the formulation of Mini Pacman from \cite{weber2017imagination}. Briefly, the agent navigates a maze to collect food while being chased by ghosts. However, we implement this task with a viewport: a $5 \times 5$ window surrounding the agent that comprises the perceptual input. The task is therefore partially observable, since the agent must navigate the space and take in information through this viewport. Thus, the agent must predict the dynamics of the ghosts \textit{in memory}, and plan its navigation accordingly, also based on remembered information about which food has already been picked up. We also point the reader to the appendix for a description and results of another RL task called BoxWorld, which demands relational reasoning in memory space.

\subsection{Language Modeling}
Finally, we investigate the task of word-based language modeling. We model the conditional probability $p(w_t | w_{<t})$ of a word $w_t$ given a sequence of observed words $w_{<t} = \left( w_{t-1}, w_{t-2}, \ldots, w_1 \right)$. Language models can be directly applied to predictive keyboard and search-phrase completion, or they can be used as components within larger systems, e.g. machine translation \cite{cho2014learning}, speech recognition \cite{bahdanau2016end}, and information retrieval \cite{hiemstra2001using}.  RNNs, and most notably LSTMs, have proven to be state-of-the-art on many competitive language modeling benchmarks such as Penn Treebank \cite{yang2017breaking, marcus1993building}, WikiText-103 \cite{rae2018fast, merity2016pointer}, and the One Billion Word Benchmark \cite{jozefowicz2016exploring, chelba2013one}. As a sequential reasoning task, language modeling allows us to assess the RMC's ability to process information over time on a large quantity of natural data, and compare it to well-tuned models. 

We focus on datasets with contiguous sentences and a moderately large amount of data. WikiText-103 satisfies this set of requirements as it consists of Wikipedia articles shuffled at the article level with roughly $100M$ training tokens, as do two stylistically different sources of text data: books from Project Gutenberg\footnote{Project Gutenberg. (n.d.). Retrieved January 2, 2018, from www.gutenberg.org} and news articles from GigaWord v5 \cite{parker2011english}. Using the same processing from \cite{rae2018fast} these datasets consist of $180M$ training tokens and $4B$ training tokens respectively, thus they cover a range of styles and corpus sizes. We choose a similar vocabulary size for all three datasets of approximately $250,000$, which is large enough to include rare words and numeric values.
\section{Results}

\subsection{$N^{th}$ Farthest}
This task revealed a stark difference between our LSTM and DNC baselines and RMC when training on $16$-dimensional vector inputs. Both LSTM and DNC models failing to surpass $30\%$ best batch accuracy and the RMC consistently achieving $91\%$ at the end of training (see figure \ref{fig:lstm_dnc} in the appendix for training curves). The RMC achieved similar performance when the difficulty of the task was increased by using $32$-dimensional vectors, placing a greater demand on high-fidelity memory storage. However, this performance was less robust with only a small number of seeds/model configurations demonstrating this performance, in contrast to the $16$-dimensional vector case where most model configurations succeeded. 

\begin{figure}
    \centering
    \includegraphics[width=1\textwidth]{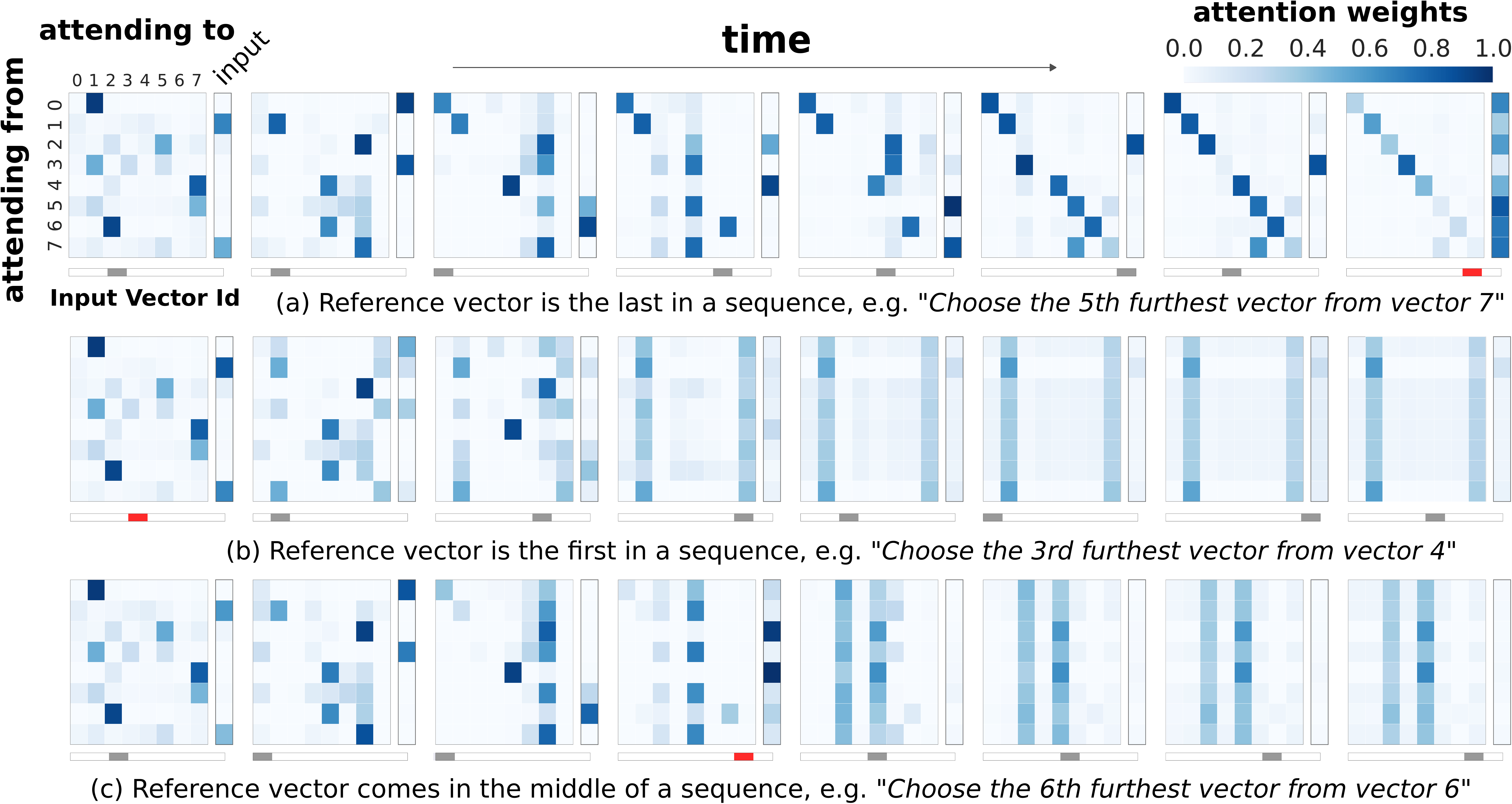}
    \caption{\textbf{Model analysis}. Each row depicts the attention matrix at each timestep of a particular sequence. The text beneath spells out the particular task for the sequence, which was encoded and provided to the model as an input. We mark in red the vector that is referenced in the task: e.g., if the model is to choose the $2^{nd}$ farthest vector from vector $7$, then the time point at which vector $7$ was input to the model is depicted in red. A single attention matrix shows the attention weights from one particular memory slot (y-axis) to another memory slot (columns), or the input (offset column), with the numbers denoting the memory slot and ``input'' denoting the input embedding.}
    \label{fig:analysis}
\end{figure}

An attention analysis revealed some notable features of the RMC's internal functions. Figure \ref{fig:analysis} shows attention weights in the RMC's memory throughout a sequence: the first row contains a sequence where the reference vector $m$ was observed last; in the second row it was observed first; and in the last row it was observed in the middle of the sequence. Before $m$ is seen the model seems to shuttle input information into one or two memory slots, as shown by the high attention weights from these slots' queries to the input key. After $m$ is seen, most evident in row three of the figure, the model tends to change its attention behaviour, with all the memory slots preferentially focusing attention on those particular memories to which the $m$ was written. Although this attention analysis provides some useful insights, the conclusions we can make are limited since even after a single round of attention the memory can become highly distributed, making any interpretations about information compartmentalisation potentially inaccurate.

\subsection{Program Evaluation}

Program evaluation performance was assessed via the \textit{Learning to Execute} tasks \cite{zaremba2014lte}. We evaluated a number of baselines alongside the RMC including an LSTM \cite{hochreiter1998lstm, pascanu2013dlstm}, DNC \cite{graves2016hybrid}, and a bank of LSTMs resembling Recurrent Entity Networks \cite{henaff2017entnet} (EntNet) - the configurations for each of these is described in the appendix.  Best test batch accuracy results are shown in Table \ref{tab:lte}. The RMC performs at least as well as all of the baselines on each task. It is marginally surpassed by a small fraction of performance on the \textit{double} memorization task, but both models effectively solve this task. Further, the results of the RMC outperform all equivalent tasks from \cite{zaremba2014lte} which use teacher forcing even when evaluating model performance. It's worth noting that we observed better results when we trained in a non-auto-regressive fashion - that is, with no teacher forcing during training. This is likely related to the effect that relaxing the ground truth requirement has on improving model generalization \cite{sbengio2015schedsamp} and hence, performance. It is perhaps more pronounced in these tasks due to the independence of output token probabilities and also the sharply uni-modal nature of the output distribution (that is, there is no ambiguity in the answer given the program).

\begin{table}[h!]
    \centering
    \caption{Test per character Accuracy on Program Evaluation and Memorization tasks. }
    \begin{tabular}{l c c c c c c}
    \toprule
    \textbf{Model} & Add & Control & Program & Copy & Reverse & Double \\
    \midrule
    LSTM \cite{hochreiter1998lstm, pascanu2013dlstm} & 99.8 & 97.4 & 66.1 & 99.8 & 99.7 & 99.7 \\
    EntNet \cite{henaff2017entnet} & 98.4 & 98.0 & 73.4 & 91.8 & \textbf{100.0} & 62.3 \\
    DNC \cite{graves2016hybrid} & 99.4 & 83.8 & 69.5 & \textbf{100.0} & \textbf{100.0} & \textbf{100.0}  \\ 

    \midrule
    Relational Memory Core & \textbf{99.9} & \textbf{99.6} & \textbf{79.0} & \textbf{100.0} & \textbf{100.0} & 99.8 \\
    \bottomrule
    \end{tabular}
    \label{tab:lte}
\end{table}

\begin{table}[h!]
    \centering
    \caption{Validation and test perplexities on WikiText-103, Project Gutenberg, and GigaWord v5.}
    \begin{tabular}{lccccc}
    \toprule
    \multicolumn{1}{c}{} & \multicolumn{2}{c}{WikiText-103} & \multicolumn{2}{c}{Gutenberg} & \multicolumn{1}{c}{GigaWord}\\
    & Valid. & Test & Valid & Test & Test \\
    \midrule
    LSTM \cite{grave2016improving} & - & 48.7 & - & - & - \\
    Temporal CNN \cite{bai2018convolutional} & - & 45.2 & - & - & - \\
    Gated CNN \cite{dauphin2016language} & - & 37.2 & - & - & - \\ 
    LSTM \cite{rae2018fast} & 34.1 & 34.3 &  41.8 & 45.5 & 43.7\\
    Quasi-RNN \cite{merity2018scalable} & 32 & 33 & - & - & -  \\
    \midrule
    Relational Memory Core & \textbf{30.8} & \textbf{31.6} & \textbf{39.2} & \textbf{42.0}  & \textbf{38.3} \\
    \bottomrule
    \end{tabular}
    \label{tab:wiki}
\end{table}

\subsection{Mini-Pacman}
In Mini Pacman with viewport the RMC achieved approximately $100$ points more than an LSTM ($677$ vs. $550$), and when trained with the full observation the RMC nearly doubled the performance of an LSTM ($1159$ vs. $598$, figure \ref{fig:pacman}).

\subsection{Language Modeling}
For all three language modeling tasks we observe lower perplexity when using the relational memory core, with a drop of $1.4 - 5.4$ perplexity over the best published results. Although small, this constitutes a $5 - 12$\% relative improvement and appears to be consistent across tasks of varying size and style. For WikiText-103, we see this can be compared to LSTM architectures \citep{graves2016hybrid, rae2018fast}, convolutional models \citep{dauphin2016language} and hybrid recurrent-convolutional models \citep{merity2018scalable}.

The model learns with a slightly better data efficiency than an LSTM (appendix figure \ref{fig:wiki_curves}). The RMC scored highly when the number of context words provided during evaluation were relatively few, compared to an LSTM which profited much more from a larger context (supplementary figure \ref{fig:wiki_context_length}). This could be because RMC better captures short-term relations, and hence only needs a relatively small context for accurate modeling. Inspecting the perplexity broken down by word frequency in supplementary table \ref{tab:giga_word_frequency}, we see the RMC improved the modeling of frequent words, and this is where the drop in overall perplexity is obtained. 
\section{Discussion}
A number of other approaches have shown success in modeling sequential information by using a growing buffer of previous states \citep{BahdanauCB14, vaswani2017attention}. These models better capture long-distance interactions, since their computations are not biased by temporally local proximity. However, there are serious scaling issues for these models when the number of timesteps is large, or even unbounded, such as in online reinforcement learning (e.g., in the real world). Thus, some decisions need to be made regarding the size of the past-embedding buffer that should be stored, whether it should be a rolling window, how computations should be cached and propagated across time, etc. These considerations make it difficult to directly compare these approaches in these online settings. Nonetheless, we believe that a blend of purely recurrent approaches with those that scale with time could be a fruitful pursuit: perhaps the model accumulates memories losslessly for some chunk of time, then learns to compress it in a recurrent core before moving onto processing a subsequent chunk. 

We proposed intuitions for the mechanisms that may better equip a model for complex relational reasoning. Namely, by explicitly allowing memories to interact either with each other, with the input, or both via MHDPA, we demonstrated improved performance on tasks demanding relational reasoning across time. We would like to emphasize, however, that while these intuitions guided our design of the model, and while the analysis of the model in the $N^{th}$ farthest task aligned with our intuitions, we cannot necessarily make any concrete claims as to the causal influence of our design choices on the model's capacity for relational reasoning, or as to the computations taking place within the model and how they may map to traditional approaches for thinking about relational reasoning. Thus, we consider our results primarily as evidence of \textit{improved function} -- if a model can better solve tasks that require relational reasoning, then it must have an increased capacity for relational reasoning, even if we do not precisely know why it may have this increased capacity. In this light the RMC may be usefully viewed from multiple vantages, and these vantages may offer ideas for further improvements. 

Our model has multiple mechanisms for forming and allowing for interactions between memory vectors: slicing the memory matrix row-wise into slots, and column-wise into heads. Each has its own advantages (computations on slots share parameters, while having more heads and a larger memory size takes advantage of more parameters). We don't yet understand the interplay, but we note some empirical findings. First, in the the $N^{th}$ farthest task a model with a single memory slot performed better when it had more attention heads, though in all cases it performed worse than a model with many memory slots. Second, in language modeling, our model used a single memory slot. The reasons for choosing a single memory here were mainly due to the need for a large number of parameters for LM in general (hence the large size for the single memory slot), and the inability to quickly run a model with both a large number of parameters and multiple memory slots. Thus, we do not necessarily claim that a single memory slot is best for language modeling, rather, we emphasize an interesting trade-off between number of memories and individual memory size, which may be a task specific ratio that can be tuned. Moreover, in program evaluation, an intermediate solution worked well across subtasks ($4$ slots and heads), though some performed best with $1$ memory, and others with $8$. 

Altogether, our results show that explicit modeling of memory interactions improves performance in a reinforcement learning task, alongside program evaluation, comparative reasoning, and language modeling, demonstrating the value of instilling a capacity for relational reasoning in recurrent neural networks. 

\section*{Acknowledgements}
We thank Caglar Gulcehre, Matt Botvinick, Vinicius Zambaldi, Charles Blundell, S\'ebastien Racaniere, Chloe Hillier, Victoria Langston, and many others on the DeepMind team for critical feedback, discussions, and support.

\clearpage
\small
\bibliographystyle{unsrt}
\bibliography{bibliography}

\clearpage
\appendix

\section{Further task details, analyses, and model configurations}

In the following sections we provide further details on the experiments and the model configurations. We will sometimes refer to the following terms when describing the model: 
\begin{itemize}
    \item ``total units'': The total number of elements in the memory matrix $M$. Equivalent to the size of each memory multiplied by the number of memories.
    \item ``num heads'': The number of attention heads; i.e., the number of unique sets of queries, keys, and values produced for the memories. 
    \item ``memory slots'' or ``number of memories'': Equivalent to the number of rows in matrix $M$.
    \item ``num blocks'': The number of iterations of attention performed at each time-step.
    \item ``gate style'': Gating per unit or per memory slot
\end{itemize}

\subsection{$N^{th}$ Farthest}
Inputs consisted of sequences of eight randomly sampled, $16$-dimensional vectors from a uniform distribution $x_t \sim \mathcal{U}(-1, 1)$, and vector labels $l_t \sim \{1, 2, ..., 8 \}$, encoded as a one-hot vectors and sampled without replacement. Labels were \textit{sampled} and hence did not correspond to the time-points at which the vectors were presented to the model. Appended to each vector-label input was the task specification (i.e., the values of $n$ and $m$ for that sequence), also encoded as one-hot vectors. Thus, an input for time-step $t$ was a 40-dimensional vector $(x_t; l_t; n; m)$.

For all models (RMC, LSTM, DNC) we used the Adam optimiser \cite{kingma2014adam} with a batch size of $1600$, learning rates tuned between $1e^{-5}$ and $1e^{-3}$, and trained using a softmax cross entropy loss function. All the models had an equivalent 4-layer MLP ($256$ units per layer with ReLu non-linearities) to process their outputs to produce logits for the softmax. Learning rate did not seem to influence performance, so we settled on $1e^{-4}$ for the final experiments.

For the LSTM and DNC, architecture parameters seemingly made no difference to model performance. For the LSTM we tried hidden sizes ranging from $64$ up to $4096$ units, and for the DNC we tried $1$, $8$, or $16$ memories, $128$, $512$, or $1024$ memory sizes (which we tied to the controller LSTM size), and $1$, $2$, or $4$ memory reads \& writes. The DNC used a $2$-layer LSTM controller. 

\begin{figure}[h]
    \centering
    \includegraphics[width=1\textwidth]{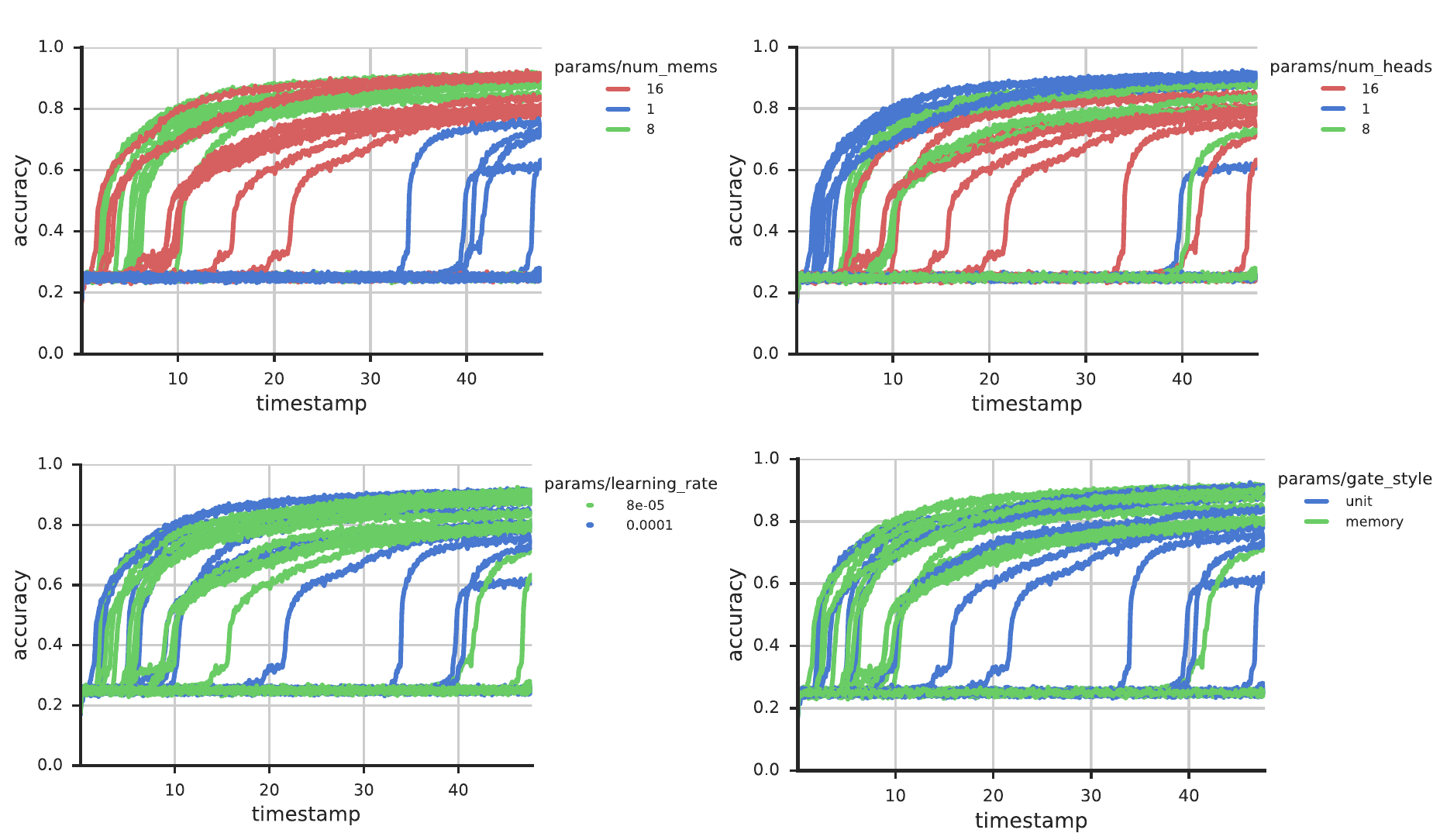}
    \caption{\textbf{$N^{th}$ Farthest hyperparameter analysis}. Timestamp refers to hours of training. There is a clear effect with the number of memories, with $8$ or $16$ memories being better than $1$. Interestingly, when the model had $1$ memory we observed an effect with the number of heads, with more heads ($8$ or $16$) being better than one, possibly indicating that the RMC can learn to compartmentalise and relate information across heads in addition to across memories.}
    \label{fig:hp_analysis}
\end{figure}

For the RMC we used $1$, $8$, or $16$ memories with $2048$ total units (so, the size of each memory was $\frac{2048}{\text{num\_mems}}$), $1$, $8$, or $16$ heads, $1$ block of attention, and both the `unit' and `memory' gating methods. Figure \ref{fig:hp_analysis} shows the results of a hyperparameter sweep scaled according to wall-clock time (models with more but smaller memories are faster to run than those with fewer but larger memories, and we chose to compare models with equivalent number of total units in the memory matrix $M$). 

\begin{figure}
    \centering
    \includegraphics[width=1\textwidth]{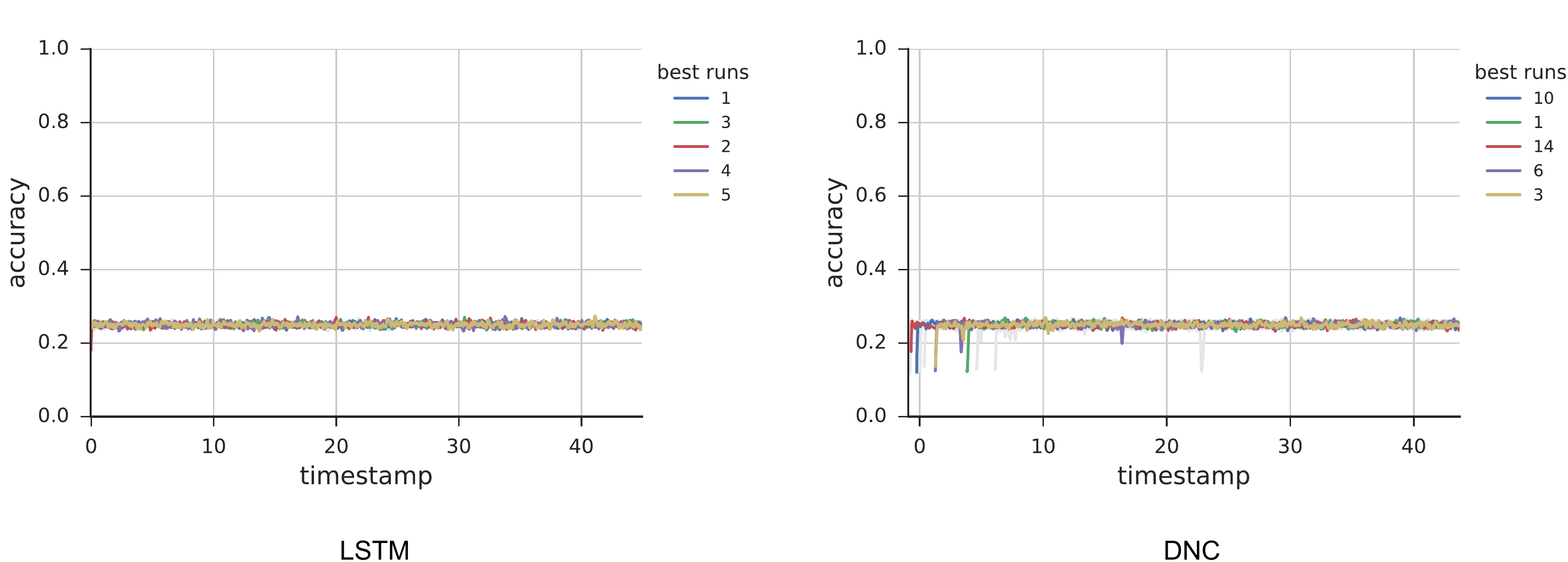}
    \caption{\textbf{LSTM and DNC training curves for the $N^{th}$ Farthest task}. }
    \label{fig:lstm_dnc}
\end{figure}

\subsection{Program Evaluation}
To further study the effect of relational structure on working memory and symbolic representation we turned to a set of problems that provided insights into the RMC's fitness as a generalized computational model.  The \textit{Learning to Execute} (\textbf{LTE}) dataset \cite{zaremba2014lte} provided a good starting point for assessing the power of our model over this class of problems.  Sample problems are of the form of linear time, constant memory, mini-programs.

\begin{figure}
    \centering
    \includegraphics[width=1\textwidth]{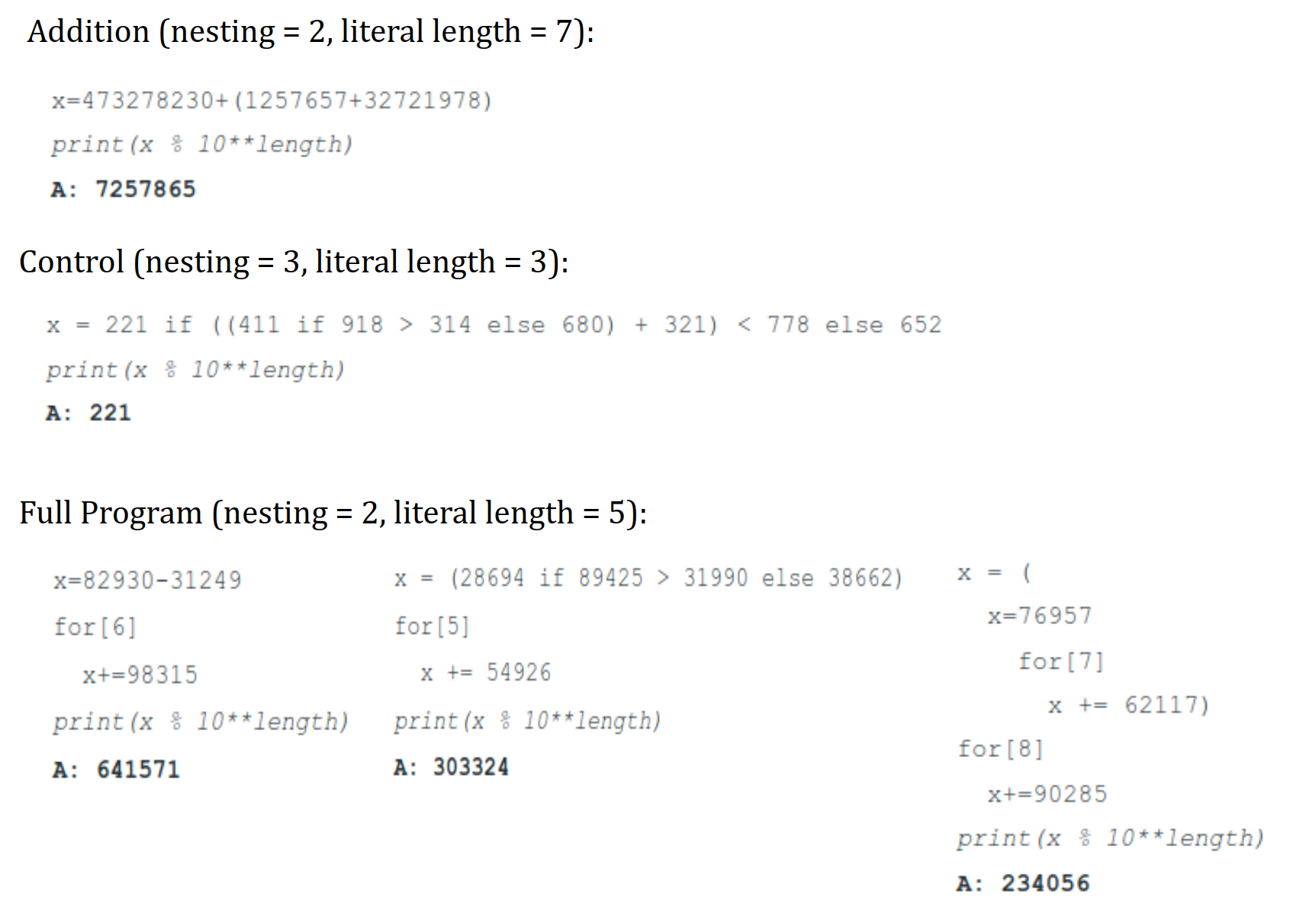}
    \caption{Samples of \textit{programmatic} tasks.  Note that training samples will sample literal length up to including the maximum length. }
    \label{fig:psamp}
\end{figure}

\begin{figure}
    \centering
    \includegraphics[width=0.7\textwidth]{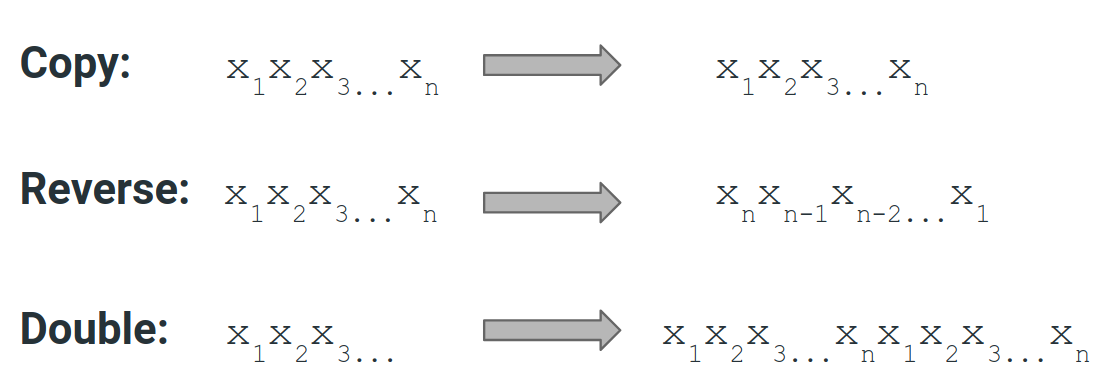}
    \caption{\textit{Memorization} tasks.  Each sub-task takes the form of a list permutation. }
    \label{fig:msamp}
\end{figure}

Training samples were generated in batches of $128$ on-the-fly. Each model was trained for $200$K iterations using an Adam optimiser and learning rate of $1e^{-3}$. The samples were parameterized by literal length and nesting depth which define the length of terminal values in the program snippets and the level of program operation nesting. Within each batch the literal length and nesting value was sampled uniformly up to the maximum value for each - this is consistent with the \textit{Mix} curriculum strategy from \cite{zaremba2014lte}. We evaluated the model against a batch of $12800$ samples using the maximum nesting and literal length values for all samples and report the top score.  Examples of samples for each task can be found in figure \ref{fig:psamp} and figure \ref{fig:msamp}.  It also worth noting that the modulus operation was applied to \textit{addition}, \textit{control}, and \textit{full program} samples so as to bound the output to the maximum literal length in case of longer for-loops.

\begin{figure}
    \centering
    \includegraphics[width=1\textwidth]{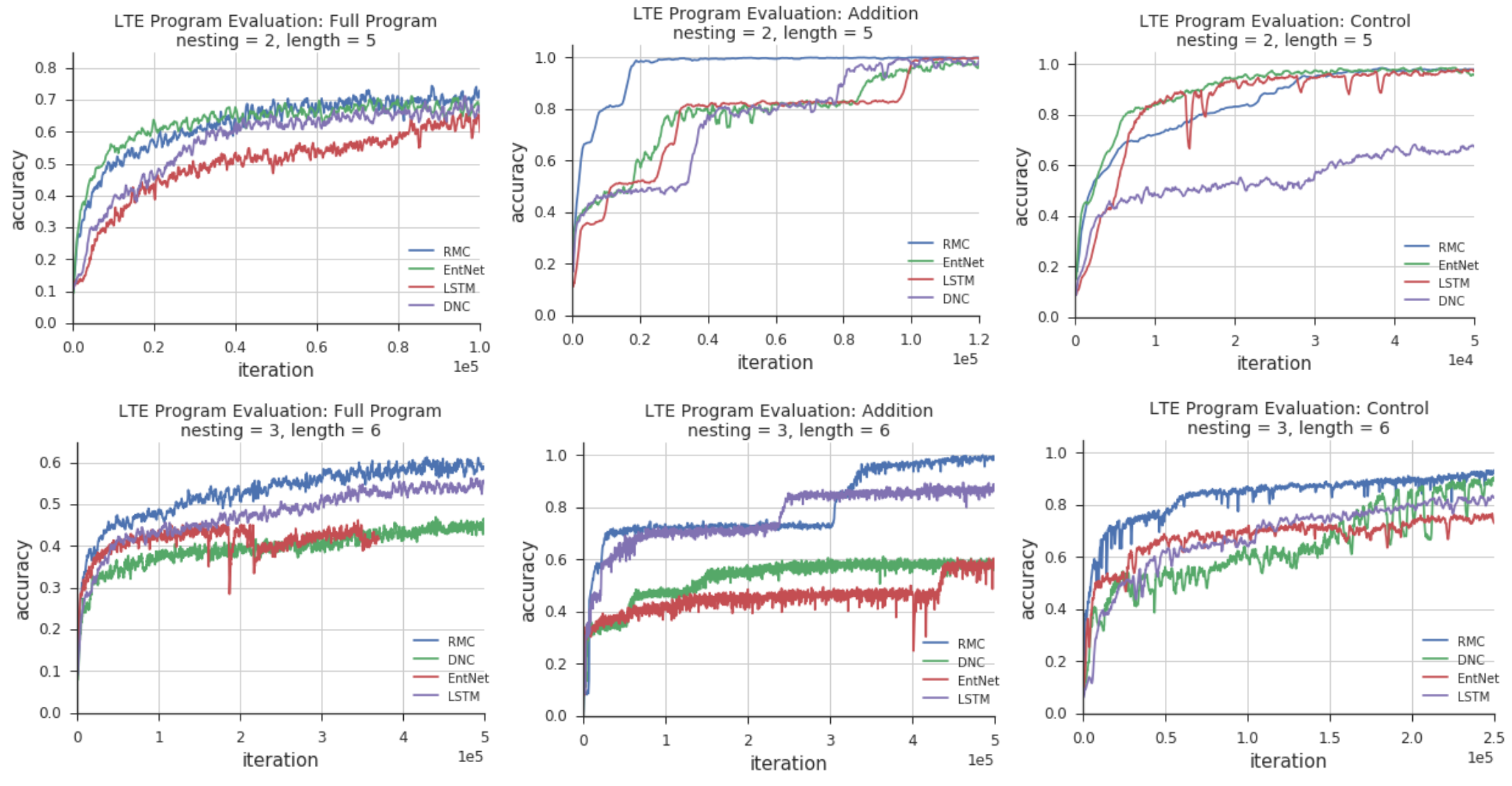}
    \caption{\textit{Programmatic} results.  From left to right: \textit{full program}, \textit{addition}, \textit{control}. The top row depicts per character accuracy scores from tasks with nesting = 2 and literal length = 5 while the bottom row shows scores from more difficult tasks with nesting = 3 and literal length = 6.}
    \label{fig:lte_results_prog}
\end{figure}

The sequential model consists of an encoder and a decoder which each take the form of a recurrent neural network \cite{sutskever2014seq2seq, zaremba2014lte}.  Once the encoder has processed the input sequence the state of the encoder is used to initialize the decoder state and subsequently to generate the target sequence (program output).  The output from all models is passed through a 4-layer MLP - all layers have size $256$ with an output ReLU - to generate an output embedding at each step of the output sequence.

In \cite{zaremba2014lte} teacher forcing is used for both training and testing in the decode phase.  For our experiments, we began by exploring teacher forcing during training but used model predictions from the previous step as input to the the decoder at the next step when evaluating the model \cite{sutskever2014seq2seq}.  We also considered the potential effect of limiting the dependency on the ground truth altogether when training the decoder \cite{sbengio2015schedsamp} and using a non-auto-regressive regime where model predictions only were used during training.  It turned out that this approach tended to yield the strongest results.

Following are the encoder/decoder configurations for a collection of memory models that performed best over all tasks. With the RMC we swept over two and four memories, and two and four attention heads, a total memory size of $1024$ and $2048$ (divided across memories), a single pass of self attention per step and scalar memory gating.  For the baselines, the LSTM is a two layer model and we swept over models with $1024$ and $2048$ units per layer, skip connections and layer-wise outputs concatenated on the final layer.  The DNC used a memory size of $80$, word size $64$, four read heads and one write head, a 2-layer controller sweeping over $128$, $256$ and $512$ latent units per layer, larger settings than this tended to hurt performance.  Also for the DNC, an LSTM controller is used for Program Evaluation problems, and feed-forward controller for memorization.  Finally, the EntNet was compared with a total memory size of either $1024$ or $2048$ with $2$, $4$, $6$, or $8$ memory cells where total memory size is divided among memories and the states of the cells are summed to produce an output.  All results reported are from the strongest performing hyper-parameter setting for the given model. 

As seen in figure \ref{fig:lte_results_prog} the RMC tends to quickly achieve high performance relative to the baselines, this demonstrates good data efficiency for these tasks especially when compared to the LSTM.  From the same figure and table \ref{tab:lte} (the results in the table depict converged accuracy scores for nesting 2 and literal length 5) it is also clear that the RMC scores well among the full set of program evaluation tasks where the DNC faltered on the \textit{control} task and the EntNet on \textit{copy} and \textit{double} tasks.  It should finally be noted that due to the RMC model size scaling with respect to total memory size over number of memories and consequently the top performing LSTM models contained many more parameters than the top performing RMC models.

\subsection{Viewport BoxWorld}
We study a variant of BoxWorld, which is a pixel-based, highly combinatorial reinforcement learning environment that demands relational reasoning-based planning, initially developed in \cite{zambaldi2018relational}. It consists of a grid of $14 \times 14$ pixels: grey pixels denote the background, lone colored pixels are keys that can be picked up, and duples of colored pixels are locks and keys, where the right pixel of the duple denotes the color of the lock (and hence the color of the key that is needed to open the lock), and the left pixel denotes the color of the key that would be obtained should the agent open the lock. The agent is denoted by a dark grey pixel, and has four actions: \textit{up}, \textit{down}, \textit{left}, \textit{right}. To make this task demand relational reasoning in a memory space, the agent only has perceptual access to a $5 \times 5$ RGB window, or viewport, appended with an extra frame denoting the color of the key currently in possession. The goal of the task is to navigate the space, observe the key-lock combinations, and then choose the correct key-lock sequence so as to eventually receive the rewarded gem, denoted by a white pixel. 

In each level there is a unique sequence of keys-lock pairs that should be traversed to reach the gem. There are a few important factors that make this task difficult: First, keys disappear once they are used. Since we include `distractor' branches (i.e., key lock paths that lead to a dead end), the agent must be able to look ahead, and reason about the appropriate path forward to the gem so as to not get stuck. Second, the location of the keys and locks are randomised, making this task completely devoid of any spatial biases. This emphasises a capacity to reason about the relations between keys and locks, in memory, based on their abstract relations, rather than based on their spatial positions. For this reason we suspect that CNN-based approaches may struggle, since their inductive biases are tied to relating things proximal in space. 

\begin{figure}
    \centering
    \includegraphics[width=.75\textwidth]{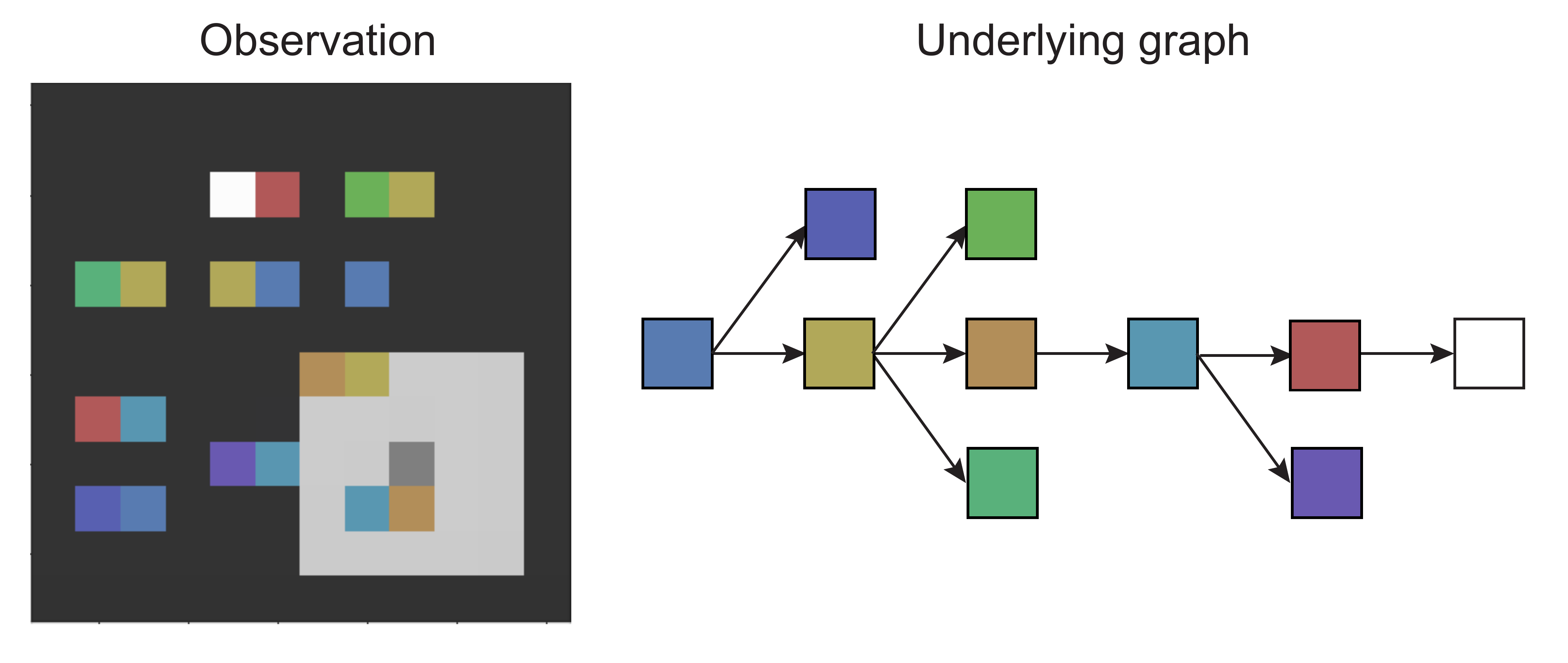}
    \caption{\textbf{Example BoxWorld level}. The left panel shows the full-view frame of a BoxWorld level. The agent, the dark grey pixel, only has access to a $5 \times 5$ view surrounding it (light gray area). The right panel shows the underlying graph that was sampled to generate the level. In this example the solution path has length 5 and there are 4 distractor branches.}
    \label{fig:boxworld}
\end{figure}

To collect a locked key the agent must be in possession of the matching key color (only one key can be held at a time) and walk over the lock, after which the lock disappears. Only then is it possible for the agent to pick up the adjacent key. Each level was procedurally generated, constrained to have only one unique sequence in each level ending with the white gem. To generate the level we first sampled a random graph (tree) that defined the possible paths that could be traversed, including distractor paths. An example path is shown in figure \ref{fig:boxworld}.

We used a total of $20$ keys and $20$ locks (i.e., colors) in our sampling pool to produce each level. Three main factors determined the difficulty of the level: (1) the path length (i.e., number of locks) to the gem; (2) the number of distractor branches; and (3) the path lengths of the distractor branches. For training we used solution path lengths of at least 1 and up to 5, ensuring that an untrained agent would have a small probability of reaching the goal by chance, at least on the easier levels. We sampled the number of distractor branches to be between 0 and 5, with a length of 1. 

The viewport observation was processed through two convolutional layers, with $12$ and $24$ kernels, and with $2 \times 2$ kernel sizes and a stride of $1$. Each layer used a ReLU non-linearity. We used two extra feature maps to tag the convolutional output with absolute spatial position ($x$ and $y$) of each pixel/cell, with the tags comprising evenly spaced values between $-1$ and $1$. The resulting stack was then passed to the RMC, containing four memories, four heads, a total memory size of $1024$ (divided across heads and memories), a single pass of self attention per step and scalar memory gating. For the baseline, we replaced the RMC with a $5 \times 5$ ConvLSTM with $64$ output channels, with $2 \times 2$ kernels and stride of 1.

We used this architecture in an actor-critic set-up, using the distributed Importance Weighted Actor-Learner Architecture \citep{espeholt2018impala}. The agent consists of $100$ actors, which generate trajectories of experience, and one learner, which directly learns a policy $\pi$ and a baseline function $V$, using the actors' experiences. The model updates were performed on GPU using mini-batches of $32$ trajectories provided by the actors via a queue. The agent had an entropy cost of $0.005$, discount ($\gamma$) of $0.99$ and unroll length of $40$ steps. The learning rate was tuned, taking values between $1\mathrm{e}{-5}$ and $2\mathrm{e}{-4}$. Informally, we note that we could replicate these results using an A3C setup, though training took longer. 

The agent received a reward of $+10$ for collecting the gem, $+1$ for opening a box in the solution path and $-1$ for opening a distractor box. The level was terminated immediately after collecting the gem or opening a distractor box.

\subsubsection{Results}
We trained an Importance Weighted Actor-Learner Architectures agent augmented with the RMC on BoxWorld levels that required opening at least 1 and up to 5 boxes. The number of distractor branches was randomly sampled from 0 to 5. This agent achieved high performance in the task, correctly solving $98\%$ of the levels after $1e9$ steps. The same agent augmented instead with a ConvLSTM performed significantly worse, reaching only $73\%$. 

\begin{figure}
    \centering
    \includegraphics[width=1\textwidth]{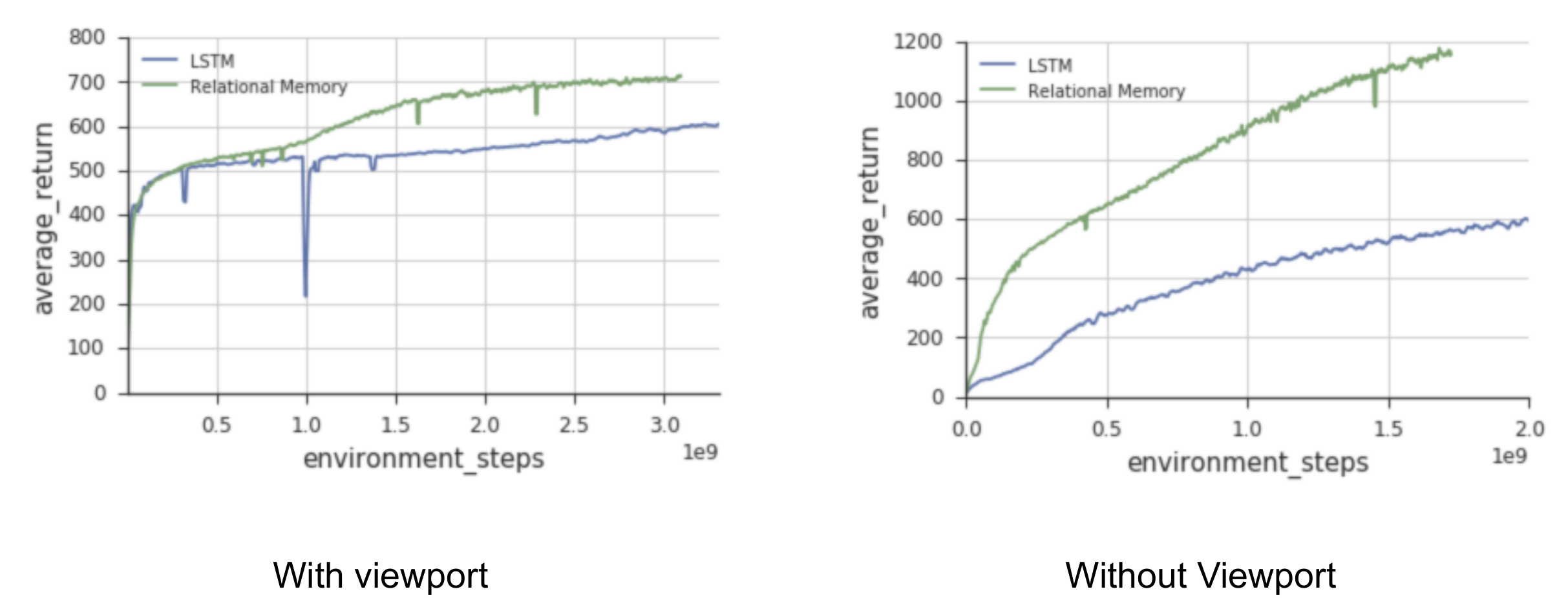}
    \caption{\textbf{Mini Pacman Results}.}
    \label{fig:pacman}
\end{figure}

\subsection{Language Modeling}
\label{app:lm}

We trained the Recurrent Memory Core with Adam, using a learning rate of $0.001$ and gradients were clipped to have a maximum L2 norm of $0.1$. Backpropagation-through-time was truncated to a window-length of $100$. The model was trained with $6$ Nvidia Tesla P100 GPUs synchronously. Each GPU trained with a batch of $64$ and so the total batch size was $384$. We used $512$ (with 0.5 dropout) as the word embedding sizes, and tied the word embedding matrix parameters to the output softmax.

We swept over the following model architecture parameters:

\begin{itemize}
    \item Total units in memory  $\{1000, 1500, 2000, 2500, 3000\}$
    \item Attention heads $\{1, 2, 3, 4, 5\}$ 
    \item Number of memories $\{1, 2\}$
    \item MLP layers $\{1, 2, 3, 4, 5\}$
    \item Attention blocks $\{1, 2, 3, 4\}$
\end{itemize}

and chose $2500$ total units, $4$ heads, $1$ memory, a $5$-layer MLP, and $1$ attention block based upon validation error on WikiText-103. We used these same parameters for GigaWord and Project Gutenberg without additional sweeps, due to the expense of training.

\begin{figure}[h!]
    \centering
    \includegraphics[width=.5\textwidth]{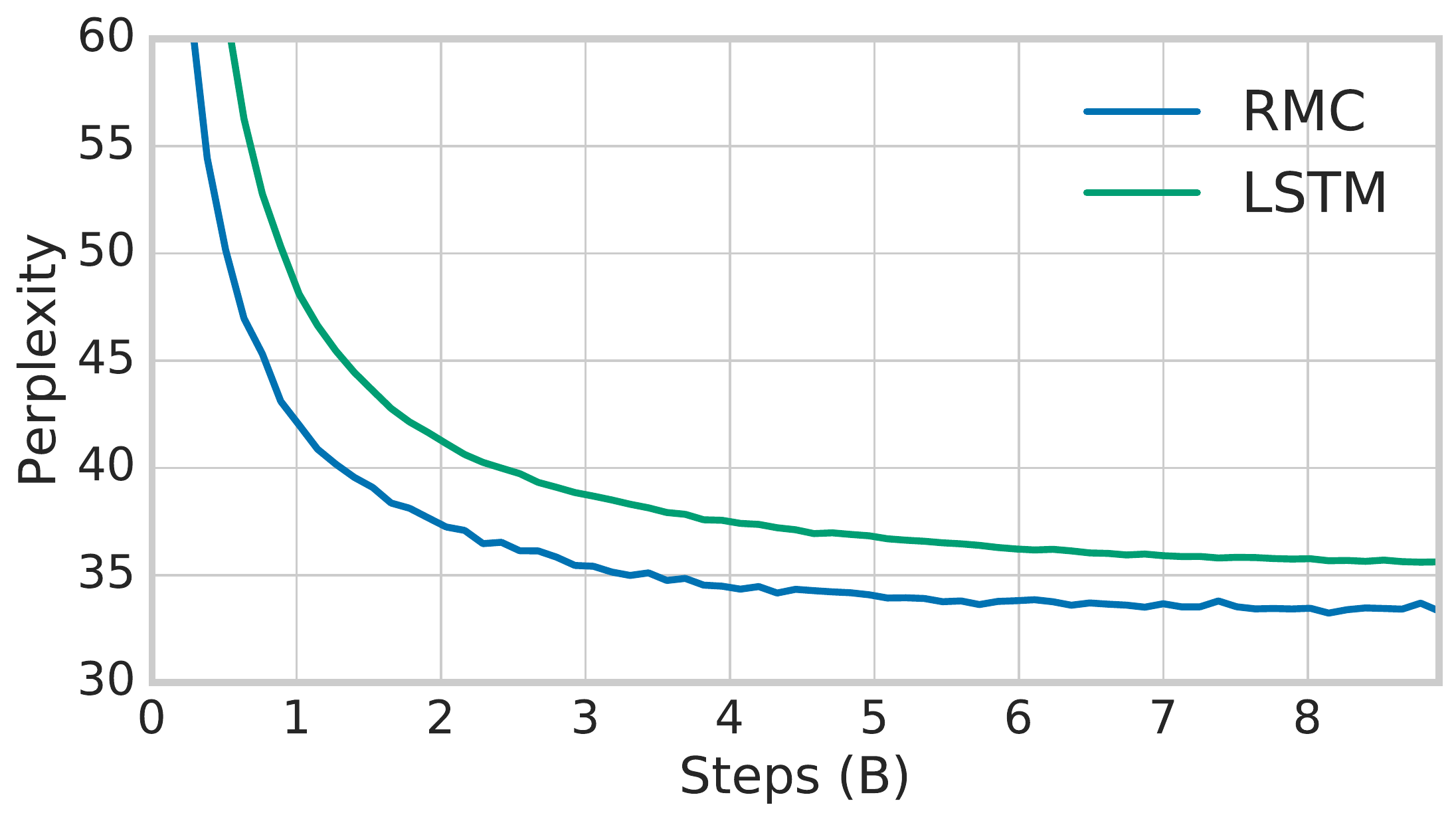}
    \caption{\textbf{Validation perplexity on WikiText-103}. LSTM comparison from \cite{rae2018fast}. Visual display of data may not match numbers from table \ref{tab:wiki} because of curve smoothing.}
    \label{fig:wiki_curves}
\end{figure}

\begin{figure}[h!]
    \centering
    \includegraphics[width=.5\textwidth]{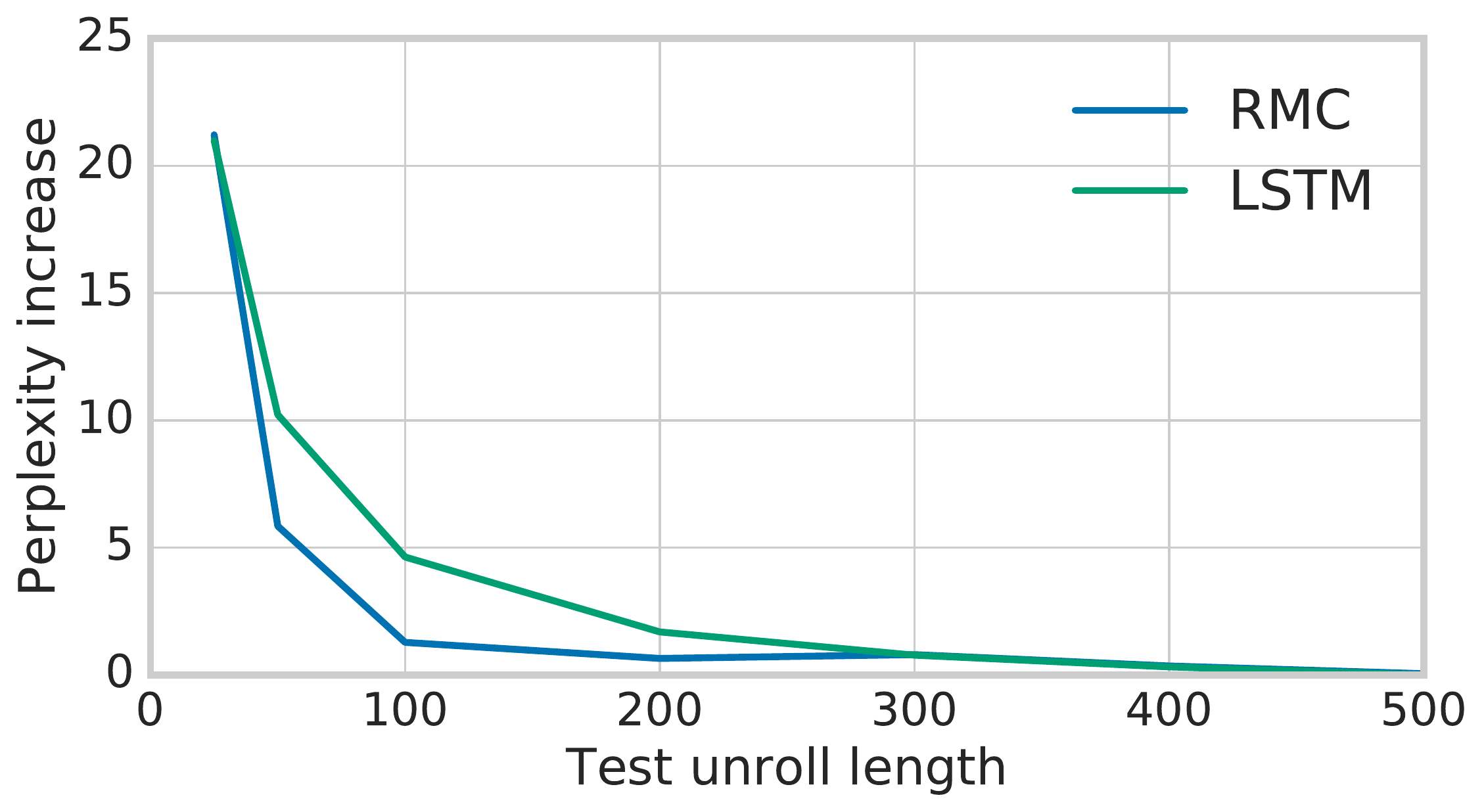}
    \caption{\textbf{Perplexity as a function of test unroll length.} Increase in perplexity when models are unrolled for shorter sequence lengths at test time without state transfer between unrolls. Perplexities are compared against the `best' perplexity where the model is unrolled continuously over the full test set. We see that both models incorporate little information beyond $500$ words. Furthermore, the RMC has a smaller gain in perplexity (drop in performance) when unrolled over shorter time steps in comparison to the LSTM, e.g. a regression of $1$ perplexity for the RMC vs $5$ for the LSTM at $100$ time steps. This suggests it is focusing on more recent words in the text.}
    \label{fig:wiki_context_length}
\end{figure}

\begin{table}[h!]
    \caption{\textbf{Test perplexity split by word frequency on GigaWord v5.} Words are bucketed by the number of times they occur in training set, $>10$K contains the most frequent words.}
    \centering
    \begin{tabular}{lcccc}
    \toprule
    & $>10$K & $10$K$ \mhyphen 1$K & $<1$K & All \\ 
    \midrule
    LSTM                   \cite{rae2018fast} & 39.4 & 6.5e3 & 3.7e4 & 53.5  \\ 
    LSTM + Hebbian Softmax \cite{rae2018fast} & 33.2 & 3.2e3 & \textbf{1.6e4} & 43.7 \\ 
    RMC                     & \textbf{28.3} & \textbf{3.1e3} & 6.9e4 & \textbf{38.3} \\
    \bottomrule
    \end{tabular}
    \label{tab:giga_word_frequency}
\end{table}

\end{document}